\documentclass[letterpaper]{article}
\usepackage{aaai19}
\usepackage{times}
\usepackage{helvet}
\usepackage{courier}
\usepackage[english]{babel}
\usepackage[utf8x]{inputenc}
\usepackage{amsmath}
\usepackage{amsfonts}
\usepackage{graphicx}
\usepackage{listings}
\usepackage{textcomp}
\usepackage{wrapfig}
\usepackage[dvipsnames]{xcolor}
\usepackage[hyphens]{url}
\usepackage[colorinlistoftodos]{todonotes}
\usepackage{microtype}
\usepackage{multirow}
\usepackage{algorithm}
\usepackage{algorithmic}
\DeclareMathOperator*{\argmax}{\arg\!\max}
\begin{document}
\title{Exploiting Class Learnability in Noisy Data}
\author{Matthew Klawonn\\ Rensselaer Polytechnic Institute \\ Dept. of Computer Science \\ Troy, NY 12180 \\ klawom@rpi.edu \And Eric Heim \\ Air Force Research Laboratory \\ Information Directorate \\ Rome, NY 13441 \\ eric.heim.1@us.af.mil \And James Hendler \\ Rensselaer Polytechnic Institute \\ Dept. of Computer Science \\ Troy, NY 12180 \\ hendler@cs.rpi.edu}
\maketitle
\begin{abstract}
In many domains, collecting sufficient labeled training data for supervised machine learning requires easily accessible but noisy sources, such as crowdsourcing services or tagged Web data. Noisy labels occur frequently in data sets harvested via these means, sometimes resulting in entire classes of data on which learned classifiers generalize poorly. For real world applications, we argue that it can be beneficial to avoid training on such classes entirely. In this work, we aim to explore the classes in a given data set, and guide supervised training to spend time on a class proportional to its \emph{learnability}. By focusing the training process, we aim to improve model generalization on classes with a strong signal. To that end, we develop an online algorithm that works in conjunction with classifier and training algorithm, iteratively selecting training data for the classifier based on how well it appears to generalize on each class. Testing our approach on a variety of data sets, we show our algorithm learns to focus on classes for which the model has low generalization error relative to strong baselines, yielding a classifier with good performance on learnable classes.
\end{abstract}

Many state of the art machine learning models require large amounts of data to avoid overfitting. In constructing large scale data sets, one often turns to accessible sources that can quickly provide vast amounts of data, including \emph{crowdsourcing} and \emph{Webly supervised learning} (scraping data and labels from the Web). Such processes frequently result in data that is noisy. For example, consider a data set of images and tags scraped from Flickr, intended to be used as training data for an image tagger. Tags on Flickr are assigned by users of the site without restriction, and the assignment of tags can range from the very literal to the abstract (see Figure \ref{fig1} for an example). When attempting to learn an image tagger, training on noisy classes is potentially harmful to model performance on classes with strong signal. Rather than try to achieve moderate performance on all classes at the expense of those with strong signal, the classifier designer may want to avoid noisy classes entirely. With this in mind, in order to develop an accurate classifier in the presence of noisy classes, we hypothesize that not all classes ought to be treated equally during training. Rather, it may be beneficial to focus training on classes for which the model appears capable of achieving a low generalization error. We call this characteristic of classes appearing in the data set \emph{learnability}.

{
\centering
\begin{figure}
\includegraphics[]{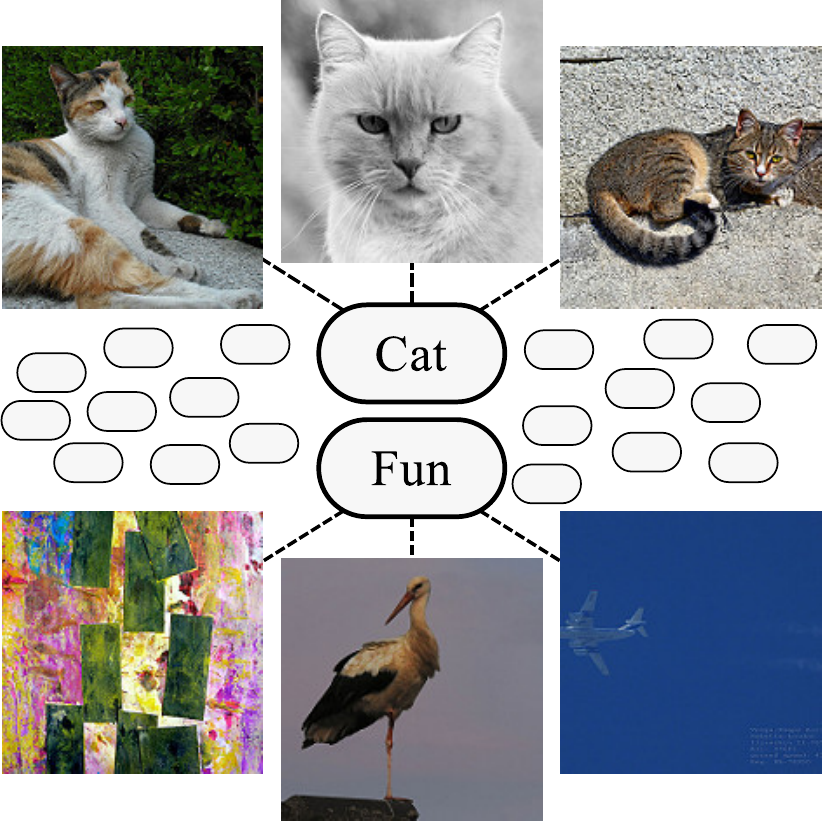}

\caption{Example of labeled Flickr images. The common features of the ``Cat" images suggest that a low generalization error can be achieved for this class, and thus that classifier training may benefit from spending more time on ``Cat'' images than ``Fun'' images.}
\label{fig1}
\end{figure}
}

To this end, we seek to develop an algorithm that uses the concept of learnability to focus the training of a classifier on specific classes. 
Our expectation is that training the model only on instances from learnable classes will yield a more accurate classifier for classes with a strong relationship between instance and label. In this work we propose an \emph{online} algorithm that acts in concert with the stochastic learning algorithm used to train a classifier.
Our algorithm assesses which of the classes are most effective in training the model, and then selects instances from those classes to use in update steps for the classifier. We pose this learning scheme as a  \emph{Multi-Armed Bandit} (MAB, bandit) problem.
By doing so, we are able to leverage prior MAB work that has developed algorithms with theoretical guarantees on performance. Our MAB algorithm is able to assess the learnability of each class by sampling only from a single class at each time-step. As a result, we are able to efficiently select classes from which to learn, incurring only slight overhead to the normal stochastic training process.

In summary, the contributions of this work are as follows:
\begin{itemize}
\item We formalize the problem of selecting learnable classes for a given model and training algorithm.
\item We develop a novel online framework that uses existing MAB algorithms to tackle the problem of selecting classes.
\item We demonstrate the effectiveness of our algorithm on both synthetic and real-world training data.
\end{itemize}

\section{Related Work}
This work is concerned with the topics of data collection from noisy sources, learning what data to learn from, and Multi-Armed Bandit approaches. We present relevant work from each of these areas.
\subsection{Data Collection}
Collecting sufficient data for training classifiers is a classic problem in machine learning. The Web is the canonical example of a noisy source of vast amounts of data, and has provided many data sets on which to train models, e.g \cite{deng2009imagenet,Philbin07,fei2007learning}. When labeling examples in these data sets, techniques like active learning \cite{settles2012active} can help limit the labeling that needs to be done by humans, only asking for labels on inputs a model cannot predict confidently. Our work is similar to active learning in that we have an algorithm guiding the selection of training data, but differs in the criteria for data selection, and we assume all data has already been labeled by some process. 

Techniques that handle noisy data with existing labels range from hand curating the data set \cite{WelinderEtal2010}, to adding model machinery intended to deal with noise \cite{sukhbaatar2014training}, to including examples based on a dynamically calculated majority vote from a crowd~\cite{deng2009imagenet}. Most standard noise-handling techniques work at the instance level.
Those that work at the label level \cite{WelinderEtal2010,garrigues2016tag} usually remove data from a training set without explicitly being informed by the training process, but rather characteristics of the data (e.g. not having sufficient quality examples). Another approach is to create new categories that account for some noise \cite{torresani2010efficient,li2013harvesting}. Works like \cite{deng2009imagenet,krishnavisualgenome} keep track of how difficult a label appears based on crowd agreement~\cite{snow2008cheap} to include or exclude various examples.
In contrast, our work attempts to automatically determine good labels based on model performance, not data set size or crowd factors.

\subsection{Learning What to Learn}
Previous methods that oversee the data selection process for training classifiers have had a variety of goals. Many models are interested in minimizing time to convergence. \cite{graves2017automated} present a curriculum learner that generates an ordering of training tasks based on task difficulty, the intuition being that easy examples help models earlier in training, while harder examples are more appropriate later. Our work follows this same core idea, yet the problem we aim to solve is fundamentally different. While they train a model to complete a pre-determined set of tasks, we learn to focus on tasks, i.e classes, that are easiest to learn for a model. Like us, \cite{ruder2017learning} are interested in manipulating the selection of training examples, though their goal is to choose examples based on suitability for transfer learning. In contrast our approach is more exploratory, as the only criteria for selecting classes is the resultant performance of a given model. Finally, \cite{fan2017learning} propose to supervise model training with a deep reinforcement learning algorithm. However, rather than actively select batches of training data, they train a filter to ignore certain examples within a given mini-batch at each training step.

\subsection{Multi-Armed Bandits}
In our approach, data is selected at each training step using a Multi-Armed Bandit algorithm, a problem first introduced in \cite{robbins1952some}. The original formulation of the problem assumes that the reward for a given decision made by the algorithm is drawn from a fixed distribution every time that decision is made (i.e. a stochastic payoff function). In our setting, rewards are based on the state of a classifier at a given training step, which changes over time. Bandit algorithms developed for such scenarios are called \emph{adversarial} MAB algorithms \cite{auer1995gambling}, and make no assumptions on the payoff structure of decisions. We additionally assume that the class set we are choosing from is large, rendering naive selection strategies ineffective. For these settings, it is common to assume a structure on the decision-space, specifically that there exists a means of measuring similarity between decisions \cite{kleinberg2013bandits}. One such approach is to assume that the reward function changes smoothly over the decision space, i.e that similar decisions will have similar payoffs \cite{Srinivas:2010:GPO:3104322.3104451}. We use a time-varying version of this algorithm presented in \cite{bogunovic2016time}.

\section{Exploiting Learnable Classes}
Given an untrained model and a labeled data set, our work aims to develop an online algorithm that identifies the learnability of classes in the data set. In determining class learnability, we aim both to 1) avoid training on difficult/noisy classes thereby yielding a strong performance over a subset of data, and 2) to make an explicit ranking of the classes based on learnability. This ranking can be used to inform downstream processes of the model's capabilities. In this section, we make precise the notion of a class learnability, present our online bandit algorithm, and develop a means of ranking classes in a data set. First, we begin with some preliminaries of Multi-Armed Bandit algorithms.

\subsection{MAB Preliminaries}
Given a collection of potential decisions $\mathcal{A}$ (called \emph{arms} in MAB nomenclature) and a number of rounds $T$ to choose (``pull") an arm, an MAB algorithm selects a sequence of arms $a_1 \ldots a_T, a_t \in \mathcal{A}$. After each round, the algorithm observes a reward $r_t(a_t)$ associated with the arm choice $a_t$. The goal of a bandit algorithm is to maximize $\sum_{i=1}^{T} r_t$, or equivalently to minimize cumulative regret. At each round $t$, regret is defined as the difference between the reward for the best arm $r^*_t := \sup_{a \in A} r_t(a)$ and the reward for the arm that was picked $r_t(a_t)$. The cumulative regret $R(T)$ is defined as the sum of regret at each round: $R(T) := \sum_{t=1}^{T} r^*_t - r_t(a_t)$. Bandit algorithms are designed to minimize $R(T)$. The challenge lies in not knowing the distributions generating $r_t(a_t) \quad \forall a \in \mathcal{A}$. In order to balance exploring promising arms with exploiting historically high reward arms, \cite{auer2002using} propose picking arms with high \emph{upper confidence bounds} (UCB). The UCB formula is given by
$$UCB(a) := \mu(a) + \beta^{\frac{1}{2}} \sigma(a)$$
with $\mu(a)$ being the mean reward of arm $a$, $\sigma(a)$ being the standard deviation of rewards of $a$, and $\beta$ being a hyperparameter that weights the exploration term. Intuitively, when an arm is largely unexplored, the uncertainty term $\beta^{\frac{1}{2}} \sigma(a)$ will dominate the UCB, leading to an exploratory arm pull. In contrast, a high average reward leads to exploitation. The balance of exploration vs exploitation can be viewed as balancing learning a reward function with using said function to make decisions. Algorithms focused only on exploitation will never learn the reward function, while algorithms focus only on exploration may not make rewarding decisions.

Finding how learnable a class is for a given model in an online fashion fits naturally into the MAB problem setting. Our proposed method proceeds in conjunction with with the stochastic training of a classifier. We treat the selection of examples during training as an arm pull, picking examples that are all labeled with a particular class, and calculate a reward for those examples intended to measure their learnability. In the next section, we formalize our MAB problem setting, define our notion of class learnability, and introduce a reward function for our MAB algorithm based on this definition.

\subsection{Problem Formulation}

In this work, we consider the \emph{multi-label} classification setting where single instances may be members of multiple classes. Let $M_\Theta$ be a multi-label classifier parameterized by $\Theta$.  Let $A$ (not to be confused with $\mathcal{A}$) be a training algorithm that iteratively updates the parameters $\Theta$ with respect to some loss function $L$. We assume that we have access to a training set $\mathcal{D} = \left\{ (x_1, y_1) \ldots (x_N, y_N)\right\}$ and a validation set $\mathcal{D}^v = \left\{ (x_1, y_1) \ldots (x_M, y_M) \right\}$. Let $C = (C_1, \ldots, C_O)$ be the set of classes reflected in the training and validation labels. We denote training and validation subsets whose instances are all labelled as a class $C_i$ by $D_{C_i}$ and $D^v_{C_i}$. 

We now formalize the notion of class learnability given $M_\Theta$, $A$, $L$, $\mathcal{D}$, and $\mathcal{D}^v$. We base our concept of learnability on model generalization. In the case of a learnable class $C_i$, we expect that $M_\Theta$ can be trained by $A$ on $D_{C_i}$ to produce a parameterization $\Theta^{*}_{C_i}$ such that $M_{\Theta^{*}_{C_i}}$ generalizes to out-of-sample instances. In our work, we measure loss on the validation set $L(\mathcal{D}^v_{C_i} | \Theta^*_{C_i})$ to approximate how well the model generalizes for $C_i$. One central assumption of this work is that certain classes in $C$ are more learnable than others, that is $L(\mathcal{D}^v_{C_i} | \Theta^*_{C_i})$ will vary over different $C_i$. This implies an ordering over classes in $C$ defined as follows:

$$C_i \preceq C_j := L(\mathcal{D}^v_{C_i} | \Theta^*_{C_i}) \leq L(\mathcal{D}^v_{C_j} | \Theta^*_{C_j})$$

For given classes $C_i \preceq C_j$, the goal of our algorithm is to guide $A$ into spending more training iterations on $C_i$ than $C_j$. We expect that doing so will not only help improve potential generalization on highly ranked classes, but also avoid negative effects that noisy classes may have on a classifier. 

In order to effectively find the above ordering on classes, it becomes important to avoid parameterizations $\Theta$ wherein $L(\mathcal{D}^v_{C_j} | \Theta) \leq L(\mathcal{D}^v_{C_i} | \Theta )$ even though $C_i \preceq C_j$, since this would lead to the class selection algorithm choosing $C_j$ more than $C_i$. Therefore we aim to train our selection algorithm \emph{online} in lockstep with $A$, receiving feedback at every iteration in order to find a good ranking as quickly as possible. In order to train our selection algorithm, we measure how the model $M_\Theta$ is progressing one step at a time, using only the information that can be gleaned from the validation subset $\mathcal{D}^v_{C_i}$.  One way to measure learning progress is via \emph{self prediction gain}~\cite{graves2017automated,oudeyer2007intrinsic}:
\begin{equation} \label{eq1}
\begin{split}
SPG(C_i) &:= L(\mathcal{D}^v_{C_i}| \Theta) - L(\mathcal{D}^v_{C_i}| \Theta')
\end{split}
\end{equation}
where $\Theta$ are the model's parameters before $A$ updates them to $\Theta'$ using $\mathcal{D}_{C_i}$.  Note that in practice, many training algorithms will use stochastically selected \emph{batches} from a subset $\mathcal{D}_{C_i}$ and/or $\mathcal{D}^v_{C_i}$, but for ease of notation we may refer to the whole set in place of a batch. Self prediction gain can be thought of as measuring how much the loss improves on the validation data after a training update. Intuitively, for a very noisy class, e.g a class with randomly assigned training and validation inputs, in expectation SPG should be zero since $M_\Theta$ is unlikely to learn anything generalizable from training examples $D_{C_i}$. For classes with more signal, it is likely that SPG will yield a positive value if the model $M_\Theta$ can be trained by $A$ to generalize. Therefore at each training step $t$, we select a class $C_t$ such that $\mathcal{D}_{C_i}$ and $\mathcal{D}^v_{C_t}$ yield a high expected value of SPG. We now outline the specific bandit algorithm used to select classes per round.

\subsection{Bandit Supervision}
At a high level, the job of the bandit algorithm at a given training step is to sample a class $C_t$ that maximizes the SPG of \eqref{eq1}. At step $t$, a history of arm pulls and their resultant SPGs are available for the bandit to approximate the true, unknown SPG function. To use this history, we require that our bandit model the uncertainty of the SPG per class; measures of uncertainty are required to balance exploration and exploitation. Further, we allow for large sets of classes $C$ (i.e arms), and assume that SPG rewards will change with time, since they depend on the changing model $M_{\Theta}$. A changing SPG function limits our ability to exploit historically successful arms. In order to handle large class sets, our bandit assumes a structure on the classes, specifically that there exists a measure of similarity between them. To handle the changing SPG function, we apply a discounting factor that limits the impact of observations made early on in training. Specifically, we utilize the bandit of \cite{bogunovic2016time}, an approach that models $f$ via a time varying \emph{Gaussian Process}.

A Gaussian Process (GP) defines a Gaussian distribution of the target function's value at each point in the function's domain. The distribution at a point $x$ is parameterized by a mean $\mu$(x) and standard deviation $\sigma(x)$, with $\mu(x)$ representing the expected value of the function at that point. GPs satisfy the need to represent the uncertainty of points in a domain via $\sigma(x)$. Further, a GP captures the relationship of a target function at different points via a covariance $\Sigma$, with the covariance between two points $x_i$ and $x_j$ given by $\Sigma(x_i,x_j) = k(x_i, x_j)$ for some positive definite kernel function $k$. Intuitively, the kernel measures similarity between two points in the domain, e.g two classes, influencing the GP to predict similar function values for similar points. By modeling the relationship between classes, we alleviate the burden of exhaustively sampling rewards for each class. In our work, we use the word2vec embedding \cite{NIPS2013_5021} of a given class (denoted $\mathbf{w}$), combined with the Matern kernel \cite{matern2013spatial} to measure similarity between classes. The Matern kernel is a generalization of the Gaussian (squared exponential) covariance function, often used because its smoothness can be controlled via a hyperparameter. Tunable smoothness is potentially valuable to our work, since natural language similarity can vary significantly depending on the domain. 

At a training step $t$, our GP defines a joint Gaussian distribution over the rewards of classes known as a \emph{prior} distribution. This prior is used to forecast the reward for selecting a training class $C_t$. After selection, the SPG for this class is calculated via \eqref{eq1}, and this value is added to the history of rewards. Our bandit updates its joint distribution by incorporating the newly observed SPG, thus producing a \emph{posterior} distribution. Calculation of the posterior is influenced by two different aspects of each entry in the SPG history: one is the inter-class similarity as defined by the kernel, the other is the time between a historical entry and the most recent observation. We now describe the selection and update steps in detail.

The selection mechanism then follows directly from the concept of upper confidence bounds and estimated priors for $\mu(x)$ and $\sigma(x)$. At time $t$, we select a class $C_t$ according to the following UCB equation

\begin{equation} \label{eq2}
\begin{split}
C_t &= \argmax_{i \in \{1, \ldots ,N\}} \mu_{t-1}(\mathbf{w}_i) + \beta^{\frac{1}{2}} \sigma_{t-1}(\mathbf{w}_i)
\end{split}
\end{equation}

where $\mathbf{w}_i$ is the word vector associated with class $C_i$. Observing the reward for the chosen class $C_t$ is as simple as calculating the loss on a validation set $\mathcal{D}^v_{C_t}$as in \eqref{eq1}. After observing a new reward, $\mu$ and $\sigma$ can be updated as in \cite{bogunovic2016time}:

\begin{equation} \label{eq3}
\begin{split}
\mu_{t}(\mathbf{w}) &:= \mathbf{k}_{t-1}(\mathbf{w})^T \left(\mathbf{K}_{t-1} + \sigma_f^2 \mathbf{I}_{t-1} \right)^{-1} \mathbf{r}_{t-1}\\
\sigma_{t}^2(\mathbf{w}) &:= k(\mathbf{w},\mathbf{w}) - \mathbf{k}_{t-1}(\mathbf{w})^T\\
& \qquad \left(\mathbf{K}_{t-1} + \sigma_f^2 \mathbf{I}_{t-1} \right)^{-1} \mathbf{k}_{t-1}(\mathbf{w})
\end{split}
\end{equation}

In order to perform this update, we have access to the history of class selections $\mathbf{C}_{t-1}$ up to time $t-1$, the corresponding vector history $\mathbf{W}_{t-1}$, and the associated rewards for these arm pulls $\mathbf{r}_{t-1}$. We then create a matrix $\mathbf{K}_{t-1}$to be $\left [ k(\mathbf{w}, \mathbf{w}') \right]_{\mathbf{w},\mathbf{w}' \in W_{t-1}} \odot \left[ (1-\epsilon)^{| i-j |/2}\right]_{i,j =1}^{t-1}$ ($\odot$ being the Hadamard product). Intuitively, this matrix compares the vectors $\mathbf{w}$ and $\mathbf{w}'$ of previously selected classes via the kernel $k$, while discounting the similarity based upon how far apart in time the two arms were pulled. The time discounting is captured via a hyperparameter $\epsilon \in [0,1]$, applied to $k$, that quantifies how much rewards are expected to change at each step; $\epsilon = 1$ indicates that rewards are independent between timesteps, and $\epsilon = 0$ means there is no variation.  $\mathbf{K}_{t-1}$ is joined with a term $\sigma_f$ that models noise produced when sampling from the underlying reward function, and is then used in conjunction with a function $\mathbf{k}_{t-1}(\mathbf{w}) := \left[ k(\mathbf{w}_i, \mathbf{w})\right]_{i=1}^{t-1} \odot \left[ (1-\epsilon)^{(t-i)/2}\right]_{i=1}^{t-1}$ that compares historical arms $\mathbf{w}_i$ to a given arm $\mathbf{w}$, again discounting old arm pulls by $\epsilon$. 

To select a class in the next round, the updated mean and variance are fed into \eqref{eq2}. When our algorithm selects a class to learn based on \eqref{eq2}, exploration will dominate if a class has not been recently selected \emph{and} no similar classes have been recently selected, since $k(\mathbf{w}, \mathbf{w})$ will surpass the other terms, yielding a high variance. Otherwise, the algorithm will select classes that the model $M_\Theta$ has done well on in the past, as $\left(\mathbf{K}_{t-1} + \sigma_f^2 \mathbf{I}_{t-1} \right)$ will become significant. We outline how this process is coupled with the training of our model $M_\Theta$ in Alg. \ref{Algorithm_1}.

\begin{algorithm}[t]
\caption{Bandit Supervised Learning. We use the default values of \protect\cite{bogunovic2016time}}
\label{Algorithm_1}
\begin{algorithmic}
\REQUIRE Exploration weighting $\beta$, time varying parameter $\epsilon$, kernel $k$
\REQUIRE Initial classifier parameters $\Theta_0$, loss function $L$, training algorithm $A$
\WHILE{$M_{\Theta}$ has not converged} 
\STATE{Sample class $C_t$ (and thus $w_t$) according to \eqref{eq2}}
\STATE{Sample training data $\mathcal{D}_{C_t}$ and validation data $\mathcal{D}^v_{C_t}$}
\STATE{$L_1 = L(\mathcal{D}^v_{C_t}, \Theta)$}
\STATE{$\Theta' \leftarrow A \left(\Theta, L(\mathcal{D}_{C_t} | \Theta) \right)$}
\STATE{$L_2 = L(X'_{c_t}, \Theta)$}
\STATE{Add $L_1-L_2$ to $\mathbf{r}$, $w_t$ to $\mathbf{W}$}
\STATE{Perform Bayesian update as in \eqref{eq3}}
\ENDWHILE
\end{algorithmic}
\end{algorithm}

\subsection{Ranking Classes}
We use class rankings to determine when bandit supervised models have converged as follows. We calculate a ranking of classes every $I$ iterations, with our experiments herein using $I=20$. We define convergence to be a sufficiently small change in ranking as measured via the \emph{normalized Kendall tau distance} \cite{kendall1955rank} given in \eqref{kendall}. This distance measures the number of pairwise differences between two rankings $\tau_{t-I}$ and $\tau_{t}$, computed at rounds $t-I$ and $t$ respectively, the set of which is given by:

\begin{equation} \label{disagreements}
\begin{split}
\mathcal{Z} &:=\{ (C_i,C_j) : i < j,\\
& (\tau_{t-I}(C_i) < \tau_{t-I}(C_j) ~ \wedge \tau_{t}(C_i) > \tau_{t}(C_j)) \\
& \vee (\tau_{t-I}(C_i) > \tau_{t-I}(C_j) ~\wedge \tau_{t}(C_i) < \tau_{t}(C_j)) \}
\end{split}
\end{equation}

where $\tau_{t}(C_i), \tau_{t-I}(C_i)$ are the ranks of class $C_i$ at rounds $t$ and $t-I$. The associated distance counts the number of differences (numerator) and normalizes by the total number of pairs (denominator).

\begin{equation} \label{kendall}
\begin{split}
Z\left(\tau_{t-I}, \tau_{t} \right) &:= \frac{\left|  \mathcal{Z} \right|}{N(N-1)/2}
\end{split}
\end{equation}
 A value of zero indicates identical rankings, while a value of one indicates rankings in the opposite order. We consider our models to have converged when $Z\left(\tau_{t-I}, \tau_{t} \right) \leq 0.05$.

There are many situations in which identifying the ordering on classes can be useful, for example in order to inform downstream processes of a classifier's capabilities. While recovering the underlying ground truth ordering $C_i \preceq C_j$ is difficult since it requires identifying an ideal set of parameters for each class, we can use the bandit history to give a good approximation, the idea being that the number of times a class was selected for training is a good indicator of the model's propensity to generalize on said class. For evaluation purposes, we are interested limiting the variability that results from the stochasticity in Alg. \ref{Algorithm_1}, and therefore produce an average ranking with information from multiple runs. This ranking is calculated by sorting the classes given the average amount each class was pulled, i.e
\begin{equation} \label{eq4}
\begin{split}
C_i \preceq C_j &:= \frac{1}{S}\sum_{s=1}^{S} P_s(C_i) \geq \frac{1}{S}\sum_{s=1}^S P_s(C_j)
\end{split}
\end{equation}
for $S$ runs of Alg. \ref{Algorithm_1} and $P_{s}(C_i)$ pulls of arm $C_i$ in run $s$. 

From a data exploration perspective, the average rankings could help in finding potentially noisy labels, or give insights into the limitations of $M_\Theta$. From an evaluation perspective, rankings can indicate how well the algorithm is working. Ideally, there would not be significant variations in the ranking list between runs, as this could indicate that $M_\Theta$ is too susceptible to small changes in initial conditions, or simply that our algorithm fails to find a proper ordering of classes. With this definition of an average ranking, we can now evaluate the quality of our bandit algorithm.

\section{Empirical Evaluation}
Loosely stated, our bandit supervision has two objectives. The first is to influence the training of a model such that the attention payed to a class reflects the learnability of said class. The second, a result of the first, is to produce a classifier that performs well on learnable classes. The goal of this section is to evaluate the performance of our bandit on these two objectives. We quantitatively evaluate performance on the first objective by using a clean data set that we pollute with label noise. For the second, we train classifiers on naturally noisier, unaltered data sets and evaluate the classification performance of our model relative to the produced class ordering. 

\subsection{Evaluating Bandit Ordering}
Given an existing data set, it is unlikely that the ordering on classes $C_i \preceq C_j$ will be known a priori. This poses a challenge from an evaluation perspective, since a ranking like that presented in \eqref{eq4} cannot be compared directly to some ground truth. We therefore create a synthetic scenario for testing our bandit algorithm. We start with the Cifar100 data set \cite{krizhevsky2009learning}, a collection of 50k training images (10k of which we set aside for validation) and 10k test images, each of which are labeled as belonging to one of a hundred classes. We then sample from training images uniformly at random, constructing 100 new subsets of and assign to each of these subsets a new, out of vocabulary label. The out of vocabulary classes are chosen from a list of the 100 most common English words that are disjoint from the original classes. In total, our newly constructed data set contains 400 training examples per original class, and 400 examples per new, noisy class.

In running our bandit algorithm on this data set, we hope to show that the produced ranking of classes favors the original classes over the new noisy classes. That is, for any original class $C_i$ and any noisy class $C_j$, we want our bandit to recover $C_i \preceq C_j$, yielding a ranking with the original classes in the top 100, and noise classes in the bottom 100. In Tab. \ref{tab:cifar}, we show the frequency with which an original class appears in the top 100 classes, ranked as in \eqref{eq4}, averaged across ten runs. Additionally, the goal of bandit supervision is to mitigate negative effects that noisy classes may have on model performance, specifically accuracy on learnable classes. As such, we compare the accuracy of a model trained without bandit supervision (ELU-N) to the same model trained with bandit supervision (MAB-S). Test set accuracy is evaluated on the original classes only, and is presented in Tab. \ref{tab:cifar}. For our classifier, we use the convolutional neural network architecture presented in \cite{clevert2015fast}, which has been shown to achieve state-of-the-art performance on Cifar100. We compare to their results that were generated by training only on the original Cifar data (ELU).

\begin {table}[t]
\begin{center}
\begin{tabular}{ |c|c|c|c| } 

\hline
 Model & Acc & R@100 & Top 10 \\
\hline 
ELU & 75.72 & -- & -- \\
\hline
ELU-N & 55.12 & -- & -- \\
\hline
MAB-S & 67.14 & 92 & Bed, Plain, Bridge,\\
& & &  Train, Cattle, Crab,\\
& & & Forest, Baby, Telephone, \\
& & & Lion \\
\hline
\end{tabular}
\end{center}
\caption {``Acc" for each model represents test-set accuracy across the original classes. ``R@100" shows the number of original classes appearing in the bandit's top 100. ``Top 10" shows the most learnable classes as decided by the bandit.} \label{tab:cifar}
\end{table}

In training the network with bandit supervision, we notice an interesting trade-off with respect to running time. The online selection mechanism prohibits time saving techniques that are common to training CNNs, such as batch pre-fetching. Therefore, each iteration of our bandit supervised training is slower than an iteration of standard mini-batch stochastic gradient descent. However, we also notice our bandit supervised network tends to converge in fewer iterations than the baselines. The aggregate result is a minimal wall-clock overhead associated with bandit supervision. 

Tab. \ref{tab:cifar} shows results in keeping with our intuition. The accuracy of our approach is superior to that of standard mini-batch selection when trained on noisy data. However, we do not fully recover the accuracy of the architecture trained on the original, noise free data. Our bandit is also able to correctly rank 92 of the original classes in the top 100. The eight noise classes ranked in the top 100, ``write", ``hot", ``word", ``water", ``call", ``sound", ``your", and ``thing", may have been chosen for a few reasons. Most likely, some are close to original classes in the word vector space, e.g ``water" is very similar to the many fish and nature related classes in Cifar100. Depending on the kernel and vector representations, sufficiently high rewards on original classes may artificially prop up a related noisy class in the eyes of the bandit. For a more naturally constructed data set, we suspect classes that are considered similar by the kernel are more likely to share similar rewards.

\subsection{Exploring Large Tag Sets}
In this section we intend to gauge the performance of bandit supervision in the face of more realistic noise. The first ``real-world" data set that we investigate is a subset of the YFCC 100M data set \cite{thomee2016yfcc100m}, a collection of 100 million tagged images and videos gathered from Flickr. In order to create a manageable but informative subset, we first gather tags of interest from the work of \cite{garrigues2016tag}, who identify tags commonly searched for by Flickr users. We then construct our data subset by gathering the first million images of YFCC 100M that are labeled with at least one of the tags in our set. In total, our data has 1 million images labeled with 4566 tags.

The classes present in Flickr data vary in learnability because they can be subjectively assigned to examples. We are also interested in learnability variation that is caused by the domain of images itself, i.e images that may be objectively and correctly labeled, yet still yield certain classes that are harder to learn than others. Recipe1m \cite{salvador2017learning} is a data set that fits this description, as it contains over 800k images paired with recipes and ingredients. We use this data to form the task of identifying ingredients given an image of a completed recipe. Many ingredients may be very difficult to identify directly, thereby producing a range of learnability across ingredients. As an example, salt is present in a significant portion of recipes across many cuisines, and yet is rarely identifiable in an image.

For these data sets, we focus on the performance of models trained by bandit supervision vs. those trained by baseline approaches. The simplest baseline to which we compare is standard mini-batch gradient descent, where batches are drawn without regard for the labels they contain. A more sophisticated baseline is to choose the target classes given the number of training examples per class, limiting training to the top $N$ most frequently occurring classes. Conventional machine learning wisdom dictates that generalizability improves as training data set size increases \cite{abu2012learning}. We hope to see the bandit select classes that do not exactly match those selected by this training-example-count-based heuristic. Instead, the bandit's selections should yield a trained model $M_\Theta$ with better quantitative performance on its chosen labels. In order to measure the performance of $M_\Theta$, we use the \emph{per class} F1-Score. Specifically, we calculate true and false positives and negatives for every class across test examples. Given the true and false positives ($TP_{C_i}$, $FP_{C_i}$) and false negatives ($FN_{C_i}$) for a class $C_i$, the recall $R_{C_i}$, precision $P_{C_i}$, and F1-Score $F_{C_i}$ are given by:

\begin{equation} \label{eq5}
\begin{split}
R_{C_i} &= \frac{TP_{C_i}}{TP_{C_i} + FN_{C_i}} \\
P_{C_i}	&= \frac{TP_{C_i}}{TP_{C_i} + FP_{C_i}} \\
F_{C_i}	&= \frac{2 \times R_{C_i} \times P_{C_i}}{R_{C_i} + P_{C_i}}
\end{split}
\end{equation}

We turn to F1-Score for evaluation because it does not rely on true negatives present in the data, and because it captures both recall and precision. In the data sets we use, true negatives far outweigh any other statistic, rendering measures like accuracy uninformative. For fairly comparing our method to baselines, we analyze the per-class performance on the top $N$ classes for each of the training methods, for various values of $N$. In training on the YFCC 100M subset, $N \in \{100, 2048\}$, while for the Recipe1m data set, $N \in \{100, 385\}$. In both sets of experiments, $M_\Theta$ is the deep convolutional architecture of \cite{garrigues2016tag}, an architecture specifically developed for tagging images with low latency during inference. Figure \ref{fig:quant} presents a quantitative analysis of the F1-Score across the top classes for each value of $N$ and the bandit, while Fig. \ref{fig:qual} qualitatively compares results between methods.

{
\centering
\begin{figure*}[t]
\includegraphics[]{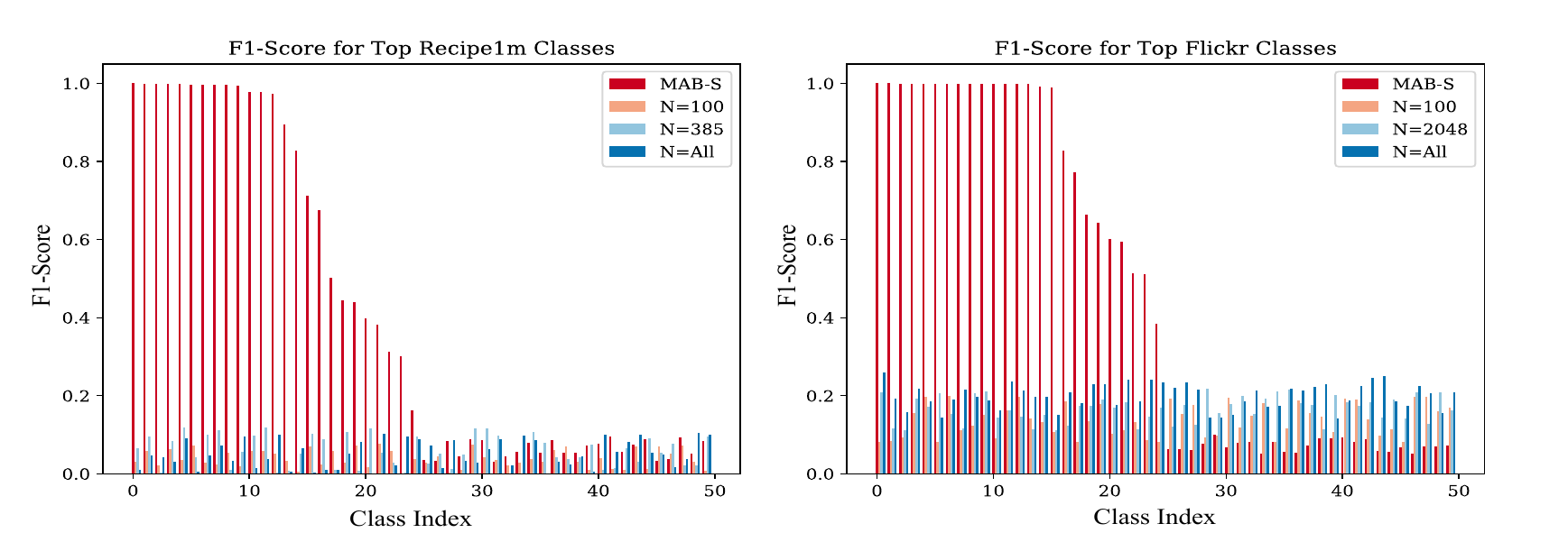}

\caption{F1-Score for the top 50 classes across methods. For bandit supervision, class ``1'' is the class most frequently chosen by the bandit. For frequency based methods, class ``1'' is the class with the highest number of occurrences.}
\label{fig:quant}
\end{figure*}
\par}

{
\centering
\begin{figure*}[t]
\includegraphics[]{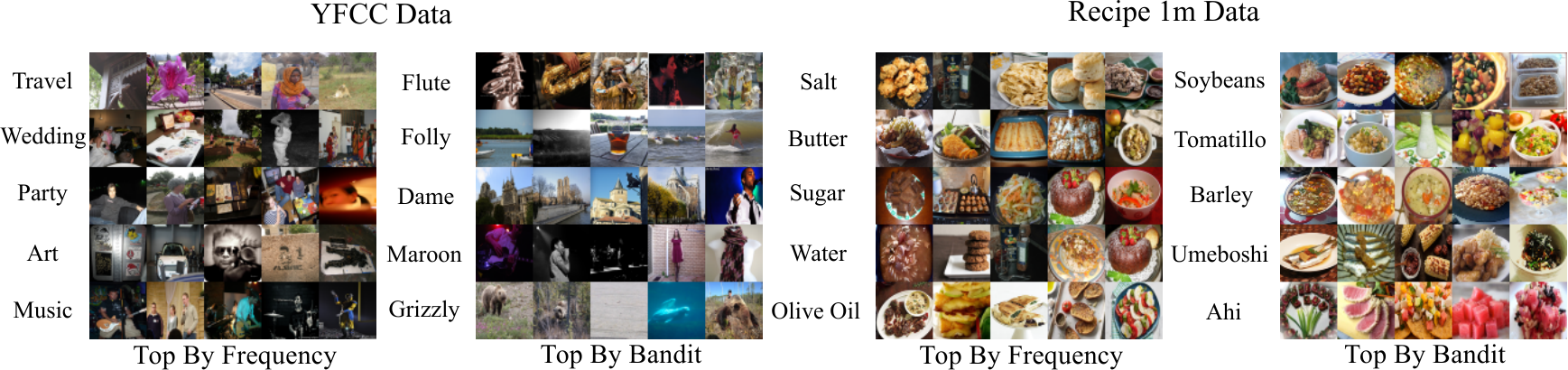}

\caption{Randomly sampled images from five classes for YFCC and Recipe1m. ``Top by Frequency" classes are the five classes with the most training data, ``Top by Bandit" are the five chosen for training most frequently by the bandit.}
\label{fig:qual}
\end{figure*}
\par}

Quantitatively, our bandit algorithm performs very well relative to the baselines for a small subset of frequently selected classes, as intended, achieving a much higher F-1 Score for top ranked classes. There is a severe dropoff after a fairly small amount of classes, though the performance of the other models suggests that both image tagging tasks are difficult. It should be noted that after the bandit-trained-model's F-1 score drops off, it generally performs worse than the other models. This simply confirms that use cases wherein performance must be above some minimum threshold for all classes are not well suited for our approach. After observing the significant divide between high and low performing classes, one question that arises is whether or not the bandit algorithm can be tuned to do well on more classes. We leave this question for future work. Qualitatively, Fig. \ref{fig:qual} shows interesting distinctions between the classes chosen by the bandit algorithm and those associated with the most training data. For the Flickr tags, we were surprised to see a concept like ``folly" ranked so highly by the bandit. Yet randomly sampled images do show similarities in this tag (3/5 displayed are images of water sports). This suggests that strong priors, i.e indications before training of what classes ought to be learnable, may be difficult to come by.

\section{Conclusion and Future Work}
In this paper, we motivated and formalized the concept of class learnability, defining it to be an ordering over classes based on potential class generalization error for a given data set. We then developed an online algorithm using existing Multi-Armed Bandit techniques that directs the training of a given classifier, with the intention of focusing on classes as their learnability dictates. We showed that this bandit algorithm is capable of identifying learnable classes in a noisy data set, and that focusing on learnable classes yields models with low generalization error relative to simple baselines and data-science inspired heuristics alike. These results were consistent across a variety of data sets, each with a unique kind of noise in the data's labeling. 

In the future, we are interested in applying our algorithm to data sets created with a variety of labeling processes. As an example, one could imagine re-labeling a data set given synonyms of existing classes, thus creating new data partitions that may facilitate better classifier performance. We believe that our bandit algorithm would be well suited to explore such a class space, identifying which concepts in a synset produce the best data partitions. Similarly, we suspect that our bandit may be appropriate for exploring fine grained recognition data sets, where there is a hierarchy on classes. In such a scenario, our bandit could act in a curriculum learning capacity, starting training by identifying coarse concepts and progressing towards more fine-grained classes if the model is ready. We also intend to investigate various other bandit formulations. As an example, there exist Combinatorial Multi-Armed Bandit algorithms that are capable of selecting a \emph{combination} of arms at a given iteration, a technique that may allow for more sophisticated batch selection and more successful optimization of a classifier.

\section{Acknowledgements}
Matthew Klawonn is sponsored by an LMI SMART scholarship. Professor Hendler’s work is supported in part by the IBM Corporation through the IBM-Rensselaer AI Research Collaboration.

\bibliography{Master.bib}

\begin{thebibliography}{}

\bibitem[\protect\citeauthoryear{Abu-Mostafa, Magdon-Ismail, and
  Lin}{2012}]{abu2012learning}
Abu-Mostafa, Y.~S.; Magdon-Ismail, M.; and Lin, H.-T.
\newblock 2012.
\newblock {\em Learning from data}, volume~4.
\newblock AMLBook New York, NY, USA:.

\bibitem[\protect\citeauthoryear{Auer \bgroup et al\mbox.\egroup
  }{1995}]{auer1995gambling}
Auer, P.; Cesa-Bianchi, N.; Freund, Y.; and Schapire, R.~E.
\newblock 1995.
\newblock Gambling in a rigged casino: The adversarial multi-armed bandit
  problem.
\newblock In {\em Foundations of Computer Science}.

\bibitem[\protect\citeauthoryear{Auer}{2002}]{auer2002using}
Auer, P.
\newblock 2002.
\newblock Using confidence bounds for exploitation-exploration trade-offs.
\newblock {\em Journal of Machine Learning Research} 3(11):397--422.

\bibitem[\protect\citeauthoryear{Bogunovic, Scarlett, and
  Cevher}{2016}]{bogunovic2016time}
Bogunovic, I.; Scarlett, J.; and Cevher, V.
\newblock 2016.
\newblock Time-varying gaussian process bandit optimization.
\newblock In {\em Artificial Intelligence and Statistics}.

\bibitem[\protect\citeauthoryear{Clevert, Unterthiner, and
  Hochreiter}{2016}]{clevert2015fast}
Clevert, D.-A.; Unterthiner, T.; and Hochreiter, S.
\newblock 2016.
\newblock Fast and accurate deep network learning by exponential linear units
  (elus).
\newblock In {\em International Conference on Learning Representations}.

\bibitem[\protect\citeauthoryear{Deng \bgroup et al\mbox.\egroup
  }{2009}]{deng2009imagenet}
Deng, J.; Dong, W.; Socher, R.; Li, L.-J.; Li, K.; and Fei-Fei, L.
\newblock 2009.
\newblock Imagenet: A large-scale hierarchical image database.
\newblock In {\em Computer Vision and Pattern Recognition}.

\bibitem[\protect\citeauthoryear{Fan \bgroup et al\mbox.\egroup
  }{2017}]{fan2017learning}
Fan, Y.; Tian, F.; Qin, T.; Bian, J.; and Liu, T.-Y.
\newblock 2017.
\newblock Learning what data to learn.
\newblock In {\em International Conference on Learning Representations
  (Workshop)}.

\bibitem[\protect\citeauthoryear{Fei-Fei, Fergus, and
  Perona}{2007}]{fei2007learning}
Fei-Fei, L.; Fergus, R.; and Perona, P.
\newblock 2007.
\newblock Learning generative visual models from few training examples: An
  incremental bayesian approach tested on 101 object categories.
\newblock {\em Computer vision and Image understanding} 106(1):59--70.

\bibitem[\protect\citeauthoryear{Garrigues \bgroup et al\mbox.\egroup
  }{2017}]{garrigues2016tag}
Garrigues, P.; Farfade, S.; Izadinia, H.; Boakye, K.; and Kalantidis, Y.
\newblock 2017.
\newblock Tag prediction at flickr: A view from the darkroom.
\newblock In {\em ACM Multimedia (Workshop)}.

\bibitem[\protect\citeauthoryear{Graves \bgroup et al\mbox.\egroup
  }{2017}]{graves2017automated}
Graves, A.; Bellemare, M.~G.; Menick, J.; Munos, R.; and Kavukcuoglu, K.
\newblock 2017.
\newblock Automated curriculum learning for neural networks.
\newblock In {\em Neural Information Processing Systems}.

\bibitem[\protect\citeauthoryear{Kendall}{1955}]{kendall1955rank}
Kendall, M.~G.
\newblock 1955.
\newblock {\em Rank Correlation Methods}, volume~2.
\newblock Hafner Publishing Co.

\bibitem[\protect\citeauthoryear{Kleinberg, Slivkins, and
  Upfal}{2013}]{kleinberg2013bandits}
Kleinberg, R.; Slivkins, A.; and Upfal, E.
\newblock 2013.
\newblock Bandits and experts in metric spaces.
\newblock {\em arXiv preprint arXiv:1312.1277}.

\bibitem[\protect\citeauthoryear{Krishna \bgroup et al\mbox.\egroup
  }{2016}]{krishnavisualgenome}
Krishna, R.; Zhu, Y.; Groth, O.; Johnson, J.; Hata, K.; Kravitz, J.; Chen, S.;
  Kalantidis, Y.; Li, L.-J.; Shamma, D.~A.; Bernstein, M.; and Fei-Fei, L.
\newblock 2016.
\newblock Visual genome: Connecting language and vision using crowdsourced
  dense image annotations.
\newblock In {\em Computer Vision and Pattern Recognition}.

\bibitem[\protect\citeauthoryear{Krizhevsky and
  Hinton}{2009}]{krizhevsky2009learning}
Krizhevsky, A., and Hinton, G.
\newblock 2009.
\newblock Learning multiple layers of features from tiny images.
\newblock Technical report, Citeseer.

\bibitem[\protect\citeauthoryear{Li, Wu, and Tu}{2013}]{li2013harvesting}
Li, Q.; Wu, J.; and Tu, Z.
\newblock 2013.
\newblock Harvesting mid-level visual concepts from large-scale internet
  images.
\newblock In {\em Computer Vision and Pattern Recognition}.

\bibitem[\protect\citeauthoryear{Mat{\'e}rn}{2013}]{matern2013spatial}
Mat{\'e}rn, B.
\newblock 2013.
\newblock {\em Spatial variation}, volume~36.
\newblock Springer Science \& Business Media.

\bibitem[\protect\citeauthoryear{Mikolov \bgroup et al\mbox.\egroup
  }{2016}]{NIPS2013_5021}
Mikolov, T.; Sutskever, I.; Chen, K.; Corrado, G.~S.; and Dean, J.
\newblock 2016.
\newblock Distributed representations of words and phrases and their
  compositionality.
\newblock In {\em Advances in Neural Information Processing Systems}.

\bibitem[\protect\citeauthoryear{Oudeyer, Kaplan, and
  Hafner}{2007}]{oudeyer2007intrinsic}
Oudeyer, P.-Y.; Kaplan, F.; and Hafner, V.
\newblock 2007.
\newblock Intrinsic motivation systems for autonomous mental development.
\newblock {\em IEEE Transactions on Evolutionary Computation} 11(6):265--286.

\bibitem[\protect\citeauthoryear{Philbin \bgroup et al\mbox.\egroup
  }{2007}]{Philbin07}
Philbin, J.; Chum, O.; Isard, M.; Sivic, J.; and Zisserman, A.
\newblock 2007.
\newblock Object retrieval with large vocabularies and fast spatial matching.
\newblock In {\em Computer Vision and Pattern Recognition}.

\bibitem[\protect\citeauthoryear{Robbins}{1952}]{robbins1952some}
Robbins, H.
\newblock 1952.
\newblock Some aspects of the sequential design of experiments.
\newblock {\em Bulletin of the American Mathematical Society} 58(5):527--535.

\bibitem[\protect\citeauthoryear{Ruder and Plank}{2017}]{ruder2017learning}
Ruder, S., and Plank, B.
\newblock 2017.
\newblock Learning to select data for transfer learning with bayesian
  optimization.
\newblock In {\em Empirical Methods in Natural Language Processing}.

\bibitem[\protect\citeauthoryear{Salvador \bgroup et al\mbox.\egroup
  }{2017}]{salvador2017learning}
Salvador, A.; Hynes, N.; Aytar, Y.; Marin, J.; Ofli, F.; Weber, I.; and
  Torralba, A.
\newblock 2017.
\newblock Recipe1m: A dataset for learning cross-modal embeddings for cooking
  recipes and food images.
\newblock {\em Transactions on Pattern Analysis and Machine Intelligence}
  720(2):619--508.

\bibitem[\protect\citeauthoryear{Settles}{2012}]{settles2012active}
Settles, B.
\newblock 2012.
\newblock Active learning.
\newblock {\em Synthesis Lectures on Artificial Intelligence and Machine
  Learning} 6(1):1--114.

\bibitem[\protect\citeauthoryear{Snow \bgroup et al\mbox.\egroup
  }{2008}]{snow2008cheap}
Snow, R.; O'Connor, B.; Jurafsky, D.; and Ng, A.~Y.
\newblock 2008.
\newblock Cheap and fast---but is it good?: Evaluating non-expert annotations
  for natural language tasks.
\newblock In {\em Empirical Methods in Natural Language Processing}.

\bibitem[\protect\citeauthoryear{Srinivas \bgroup et al\mbox.\egroup
  }{2010}]{Srinivas:2010:GPO:3104322.3104451}
Srinivas, N.; Krause, A.; Kakade, S.; and Seeger, M.
\newblock 2010.
\newblock Gaussian process optimization in the bandit setting: No regret and
  experimental design.
\newblock In {\em International Conference on Machine Learning}.

\bibitem[\protect\citeauthoryear{Sukhbaatar \bgroup et al\mbox.\egroup
  }{2014}]{sukhbaatar2014training}
Sukhbaatar, S.; Bruna, J.; Paluri, M.; Bourdev, L.; and Fergus, R.
\newblock 2014.
\newblock Training convolutional networks with noisy labels.
\newblock In {\em International Conference on Learning Representations
  (Workshop)}.

\bibitem[\protect\citeauthoryear{Thomee \bgroup et al\mbox.\egroup
  }{2016}]{thomee2016yfcc100m}
Thomee, B.; Shamma, D.~A.; Friedland, G.; Elizalde, B.; Ni, K.; Poland, D.;
  Borth, D.; and Li, L.-J.
\newblock 2016.
\newblock Yfcc100m: The new data in multimedia research.
\newblock {\em Communications of the ACM} 59(2):64--73.

\bibitem[\protect\citeauthoryear{Torresani, Szummer, and
  Fitzgibbon}{2010}]{torresani2010efficient}
Torresani, L.; Szummer, M.; and Fitzgibbon, A.
\newblock 2010.
\newblock Efficient object category recognition using classemes.
\newblock In {\em European Conference on Computer Vision}.

\bibitem[\protect\citeauthoryear{Welinder \bgroup et al\mbox.\egroup
  }{2010}]{WelinderEtal2010}
Welinder, P.; Branson, S.; Mita, T.; Wah, C.; Schroff, F.; Belongie, S.; and
  Perona, P.
\newblock 2010.
\newblock {Caltech-UCSD Birds 200}.
\newblock Technical Report CNS-TR-2010-001, California Institute of Technology.

\end{thebibliography}
\bibliographystyle{aaai}
\end{document}